\documentclass[letterpaper]{article} 
\usepackage{aaai25}  
\usepackage{times}  
\usepackage{helvet}  
\usepackage{courier}  
\usepackage[hyphens]{url}  
\usepackage{graphicx} 
\usepackage{natbib}  
\usepackage{caption} 
\usepackage{algorithm}
\usepackage{algorithmic}
\usepackage{adjustbox}
\usepackage{amssymb}
\usepackage{multirow}
\usepackage{caption}
\usepackage{fancyhdr}
\usepackage{booktabs}
\usepackage{amsmath}
\usepackage{makecell}

\usepackage[table]{xcolor} 
\usepackage{newfloat}
\usepackage{listings}
\DeclareCaptionStyle{ruled}{labelfont=normalfont,labelsep=colon,strut=off} 
\floatstyle{ruled}
\newfloat{listing}{tb}{lst}{}
\floatname{listing}{Listing}
\title{CA-Edit: Causality-Aware Condition Adapter for High-Fidelity Local Facial Attribute Editing}
\author{
    Xiaole Xian\textsuperscript{\rm 1}\equalcontrib, Xilin He\textsuperscript{\rm 1}\equalcontrib, Zenghao Niu\textsuperscript{\rm 1}, Junliang Zhang\textsuperscript{\rm 1}, Weicheng Xie\textsuperscript{\rm 1, \rm 3}\thanks{Corresponding author}, Siyang Song\textsuperscript{\rm 4}, Zitong Yu\textsuperscript{\rm 5}, Linlin Shen\textsuperscript{\rm 1, \rm 2, \rm 3}\\
}
\affiliations{
    \textsuperscript{\rm 1}Computer Vision Institute, School of Computer Science \& Software Engineering, Shenzhen University,\\
    \textsuperscript{\rm 2}National Engineering Laboratory for Big Data System Computing Technology, Shenzhen University,\\
    \textsuperscript{\rm 3}Guangdong Key Laboratory of Intelligent Information Processing,
    \textsuperscript{\rm 4}University of Exeter,
    \textsuperscript{\rm 5}Great Bay University

}

\usepackage{bibentry}

\begin{document}

\maketitle

\begin{abstract}
For efficient and high-fidelity local facial attribute editing, most existing editing methods either require additional fine-tuning for different editing effects or tend to affect beyond the editing regions. Alternatively, inpainting methods can edit the target image region while preserving external areas. However, current inpainting methods still suffer from the generation misalignment with facial attributes description and the loss of facial skin details. To address these challenges, (i) a novel data utilization strategy is introduced to construct datasets consisting of attribute-text-image triples from a data-driven perspective, (ii) a Causality-Aware Condition Adapter is proposed to enhance the contextual causality modeling of specific details, which encodes the skin details from the original image while preventing conflicts between these cues and textual conditions. In addition, a Skin Transition Frequency Guidance technique is introduced for the local modeling of contextual causality via sampling guidance driven by low-frequency alignment. Extensive quantitative and qualitative experiments demonstrate the effectiveness of our method in boosting both fidelity and editability for localized attribute editing. The code is available at \textcolor{blue}{\url{https://github.com/connorxian/CA-Edit}}.
\end{abstract}
\section{Introduction}


Efficient and high-fidelity local facial attribute editing with textual description represents a challenging task in computer vision.
GANs-based methods \cite{wang2022high, pernuvs2023maskfacegan} have explored this task, which primarily optimize the original image within the latent space 
with a pre-trained StyleGAN model \cite{karras2020analyzing}. However, these GANs-based methods require additional fine-tuning for different attributes.
Subsequently, the prior diffusion-based image editing methods based on the text-to-image (T2I) diffusion models achieve image editing in various ways.
These methods are either based on P2P \cite{hertz2022prompt}, utilizing the original image attention injection mechanism to preserve the layout, or based on DDIM Inversion \cite{song2020denoising}, modifying the latent at the noise level. However, such methods may lead to inconsistencies beyond the editing target area.
\begin{figure}[!ht]
  \centering
  \captionsetup{skip=5pt}
  \includegraphics[width=1.0\linewidth]{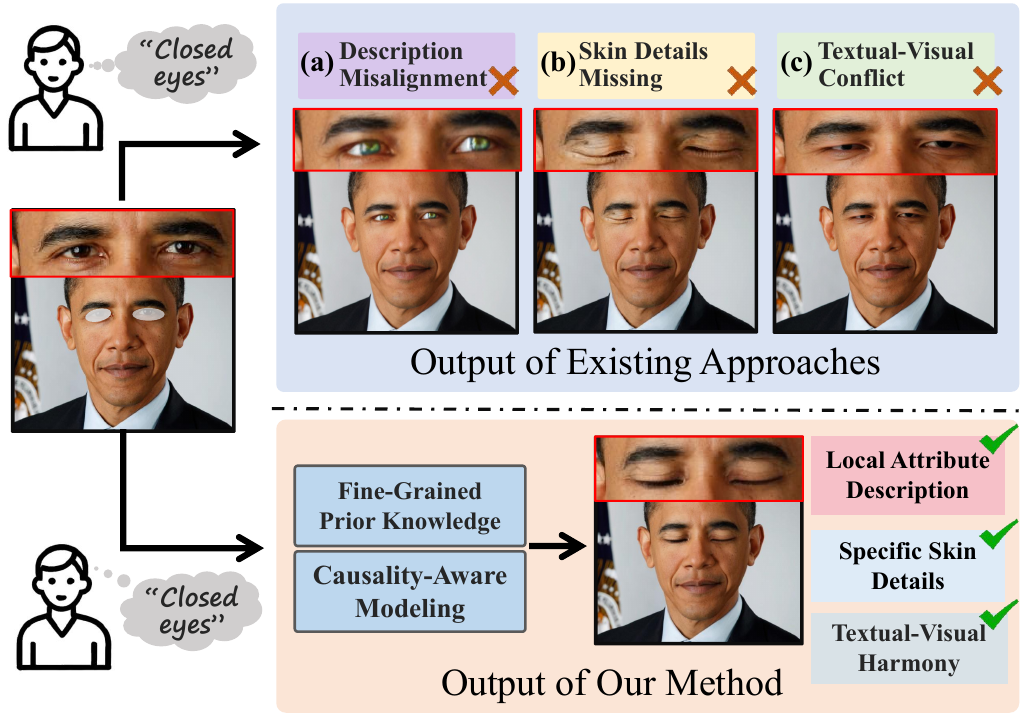}
  \caption{ \textbf{(Top)} The existing text-guided inpainting pipeline for our local attribute editing task. \textbf{(Bottom)} Our method takes account of the causality of the the specific details from the original image, improving the editability and the fidelity.}
  \label{fig:motivation}
\end{figure}
Regarding local facial attribute editing, image inpainting is a technique focused on local masked region painting, which also benefits from the recent advances in diffusion models \cite{avrahami2022blended, yang2023paint, yang2023uni}.
Besides, image inpainting has been also developed for local facial attribute editing, which focuses on the inpainting of local masked regions, based on advanced diffusion models \cite{avrahami2022blended, yang2023paint, yang2023uni}.
Text-guided image inpainting\cite{avrahami2022blended} allows prompt-driven content generation in specific areas without finetuning during inference, while maintaining consistency between the editing and unmasked regions, which is thus used in our method.

However, existing methods for image inpainting may suffer from concerns in terms of editability and fidelity.
\textbf{The first problem:} they \cite{zhang2023adding, ju2024brushnet} struggle to understand the contextual relationship between unmasked facial regions and the textual description, resulting in the neglect of the text prompt while creating a plain completion ( Fig.\ref{fig:motivation} (a) ). 
For addressing this problem, Hd-Painter \cite{manukyan2023hd} can better align the inpainting generation with the text by modifying the latent, while it still fails for local facial text prompts.
The root cause is that previous diffusion models are primarily trained on natural image-text pairs, lacking the fine-grained knowledge of human faces. 

\textbf{The second problem:} 
For facial inpainting, previous works \cite{rombach2022high, yang2023uni} do not take adequate consideration of the contextual causality between the masked region and the specific details (skin texture, skin tone, and other details) of the original image. 
The causality consideration is further constrained by the conflict between textual editing conditions and the preservation of these details in original image.
In facial images, even slight differences in these details become visibly obvious, largely impairing the overall naturalness. ( Fig.\ref{fig:motivation} (b) )
Therefore, the key to maintaining the skin details and mitigating the difference lies in the reasonably causality-aware modeling of these specific details from the original image.

For addressing this problem,
existing approaches adapt the parallel attention with textual conditions(i.e. IP-Adapter \cite{ye2023ip}) to inject original image information and enhance contextual causality modeling.
However, as shown in Fig. \ref{fig:motivation}, this causality conflicts modeling with the text condition may lead to severe content leakage. ( Fig.\ref{fig:motivation} (c) )
Meanwhile, from a localized contextual perspective, existing methods \cite{ju2024brushnet, xu2024personalized} lack explicit approaches for this fine-grained local context, causing disharmony in boundary regions of the primary editing regions, while the skin transitions are generally smooth.



To address these challenges, we proposed our CA-Edit from the local attribute data construction and causality-aware condition adapter.
\textbf{For addressing the first problem},
training on detailed textual captions of local facial attributes would be crucial for editability. To this end, we introduce a data construction pipeline, leveraging Multimodal Large Language Models (MLLMs) \cite{chen2023sharegpt4v, li2024mgm} for automatic local facial attribute captioning and the face parsing model for segmentation acquisition. 
\textbf{For addressing the second problem}, we introduce an additional adapter for original image condition, as well as a sampling guidance during inference, to fully explore original image cues. 
Specifically, (i) the Causality-Aware Condition Adapter ($\text{CA}^2$) is proposed to enhance the causality modeling while preventing the conflict with textual condition.
(ii) a sampling guidance technique called Skin Transition Frequency Guidance (STFG) is proposed to mitigate the artifacts on the `boundary regions' via enhancing the similarity between the generated image and the low-frequency components of the original image.

The main contributions of this work are summarized as:
\begin{itemize}
    \item To address the limitations of existing datasets lacking local facial attribute captions, we propose LAMask-Caption, the first dataset with detailed local facial captions which contains 200,000 high-quality facial images and employs Large Multimodal Models (MLMMs) for automatic captioning of local facial regions.
    \item To jointly address the issues of fine-grained context modeling and content leakage, we propose the novel $\text{CA}^2$) that enhances contextual causality modeling in primary editing regions while regularizing the visual condition according to the textual condition and latent.
    Furthermore, we propose the novel STFG to preserve the skin details on the boundary regions by enhancing the low-frequency similarity with the original image during inference.
    \item Quantitative and qualitative experiments demonstrate that CA-Edit produces more harmonious and natural outcomes, showcasing the superiority of our method in local attribute editing.
    
\end{itemize}

\section{Related Work}

\subsection{Generative Face Editing}
The advancement of facial editing and manipulation has been promoted by the emergence of recent generative approaches. 
Early efforts in this area have explored the application of GANs-based models \cite{karras2019style, shen2020interfacegan, yang2021discovering, xia2021tedigan}. 
MaskGAN \cite{CelebAMask-HQ} 
demonstrated the benefit of using spatially local face editing.
InterFaceGAN \cite{shen2020interfacegan} regularizes the latent code of an input image along a linear subspace. 
Recently, increasing researchers have resorted to diffusion models to enhance the generative capability for face editing.
Methods like \cite{ding2023diffusionrig, jia2023discontrolface} both explored the use of 3D modalities as reference cues to make facial image editing more robust and controllable.
Xu et al. \cite{xu2024personalized} 
finetune a diffusion model for editing tasks tailored to the individual's facial characteristics.
However, these approaches require extra conditions beyond text, limiting their suitability for our task due to user accessibility issues.

\subsection{Text-driven image editing}
Early works \cite{nitzan2022mystyle, andonian2021paint, xia2021tedigan} leveraging pretrained GAN generators \cite{karras2019style} have explored the text-driven image synthesis. 
Among approaches for semantic image editing, text-guided image editing based on diffusion models has garnered growing attention. \cite{gal2022image, ruiz2023dreambooth, rombach2022high, morelli2023ladi,mao2023guided, zhong2023adapter, brooks2023instructpix2pix} have exploited diffusion models for text-driven image editing. Textual Inversion \cite{gal2022image} generates an image by learning a concept embedding vector combined with other text features. 
For better control of the original semantic cues, InstructPix2Pix \cite{brooks2023instructpix2pix} enables image editing based on textual instructions by leveraging a conditioned diffusion model trained on a dataset generated from the combined knowledge of a language model and a text-to-image model. 
DiffusionCLIP \cite{kim2022diffusionclip} and Asyrp \cite{kwon2022diffusion} draw inspiration from GAN-based methods \cite{gal2022stylegan} that use CLIP, and use a local directional CLIP loss between image and text to manipulate images.
However, these methods either require additional finetuning or lead to changes outside target editing regions, which fail to meet the requirement of local editing.

\subsection{Diffusion Models for Inpainting}
Image inpainting is devoted to reconstructing or filling in the missing regions of an image in a visually coherent manner.
Benefited from the pretrained T2I diffusion models, many prominent works \cite{avrahami2022blended, yang2023paint, ju2024brushnet, lugmayr2022repaint, yang2023uni} that are zero-shot and do not affect the regions outside the edited area, were developed.
Stable Diffusion Inpainting \cite{rombach2022high} and ControlNet Inpainting \cite{zhang2023adding} both leverage large-scale pre-trained T2I models, fine-tune them to adapt models for this task.  During inference, the method \cite{avrahami2022blended} removes noises in a weighted manner according to the mask at each time step, which can reduce the occurrence of unnatural artifacts. \cite{levin2023differential} use a continuous mask rather than a binary mask, to enable fine-grained control over the diffusion of each pixel. Paint-by-example \cite{yang2023paint} uses image embedding to replace the original text embedding to improve image-to-image inpainting. However, due to the lack of image-text pairs of face attributes for training or adequate causality exploration in keeping the skin details, the inference stage of the aforementioned methods often results in artifacts. 

\begin{figure}[!t]
  \centering
  \captionsetup{skip=5pt}
  \includegraphics[width=1.0\linewidth]{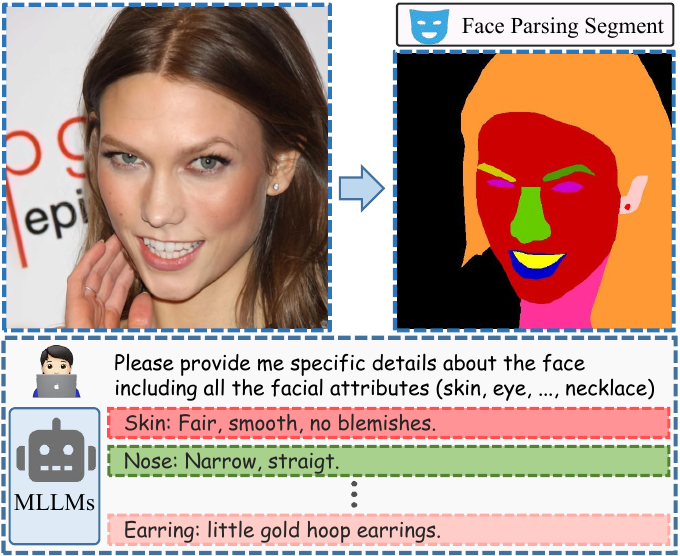}
  \caption{The pipeline of LAMask-Caption construction.}
  \label{fig:data}
\end{figure}

\section{Preliminaries}

\textbf{Diffusion Model.} 
Diffusion models are a family of generative models that consist of the processes of diffusion and denoising. The diffusion process follows the Markov chain and gradually adds Gaussian noise to the data, transforming a data sample $\mathbf{x}_{0}\sim q(\mathbf{x}_{0})$ into the noisy sample $\mathbf{x}_{1:T}=\mathbf{x}_{1},\mathbf{x}_{2},\cdots,\mathbf{x}_{T}$ in $T$ steps. The denoising process utilizes a learnable model to generate samples from this Gaussian noise distribution denoted as $p_{\theta}(\mathbf{x}_{0:T})$ at time step $t$ based on the condition $c$, where $\theta$ denotes the learnable parameters. 
Eventually, the training of the model is formulated as:
\begin{equation}
    \mathcal{L}=\mathbb{E}_{\mathbf{x}_0,\boldsymbol{\epsilon}\sim\mathcal{N}(\mathbf{0},\mathbf{I}),\boldsymbol{c},t}\|\epsilon-\epsilon_\theta(\mathbf{x}_t,\boldsymbol{c},t)\|_2^2,
\label{feature}
\end{equation}
where $\mathbf{x}_{0}$ denotes the original image, $c$, $t\in[0,T]$ represents the condition and the timestep of the diffusion process. 

\textbf{Reference Net for Diffusion Model.} 
As introduced in BrushNet\cite{brooks2023instructpix2pix} and ControlNet\cite{zhang2023adding}, a reference net is constructed by adding an additional branch dedicated to the spatial condition, which is well-suited for our task-specific mask generation.
The additional condition is first encoded with the reference net, which is then added into the skipped connections of the Stable Diffusion \cite{rombach2022high} UNet after being processed by zero convolutions. Eventually, the noise prediction of U-Net with the reference net is formulated as $\epsilon_\theta(\mathbf{x}_t, \boldsymbol{c}_{img},\boldsymbol{c}_{txt},t)$, where ${c}_{img}$ and ${c}_{txt}$ represent the image and text conditions, respectively.

\section{Method}

To enable local facial attributes inpainting, we first construct the dataset LAMask-Caption including the face images, textual descriptions of local facial attributes and the specific segmentation mask of the attributes (Fig. \ref{fig:data}). 
To adapt the T2I model to our task, we trained a reference network copied from the U-Net. Based on this network, we introduced Causality-Aware Condition Adapter ($\text{CA}^{2}$) to enhance skin detail causality while balancing textual and visual cues for precise and seamless attribute editing.
Additionally, to reduce the artifacts between generated content and the unmasked regions, our Skin Transition Frequency Guidance (STFG) technique further leverages the skin detail in the original image during inference, to avoid the effect of imprecise input masks.

\subsection{LAMask-Caption Construction Pipeline}

A key reason that current diffusion models encounter difficulties with local facial editing is the lack of precise textual captions describing local facial attributes in the training data, as mainstream diffusion models are primarily trained on large-scale natural image datasets such as Laion-2B \cite{schuhmann2022laion} or MS-COCO \cite{lin2014microsoft}. 
Hence, a face dataset with local attributes-text pairs is essential for finetuning the pretrained diffusion model to adapt to facial local attribute editing.
While the existing CelebA-dialog dataset \cite{jiang2021talkedit} and FaceCaption-15M \cite{dai202415m} contain manually annotated textual captions for each image, it mainly focuses on overall attributes (i.e. age, skin) rather than local facial attributes. Therefore, their global captions would fail to meet the demand as training data of local facial attribute editing, which motivates us to develop a new dataset with complete local facial attribute captions.

\begin{figure*}[!ht]
  \centering
  \captionsetup{skip=2pt}
  \includegraphics[width=1.0\linewidth]{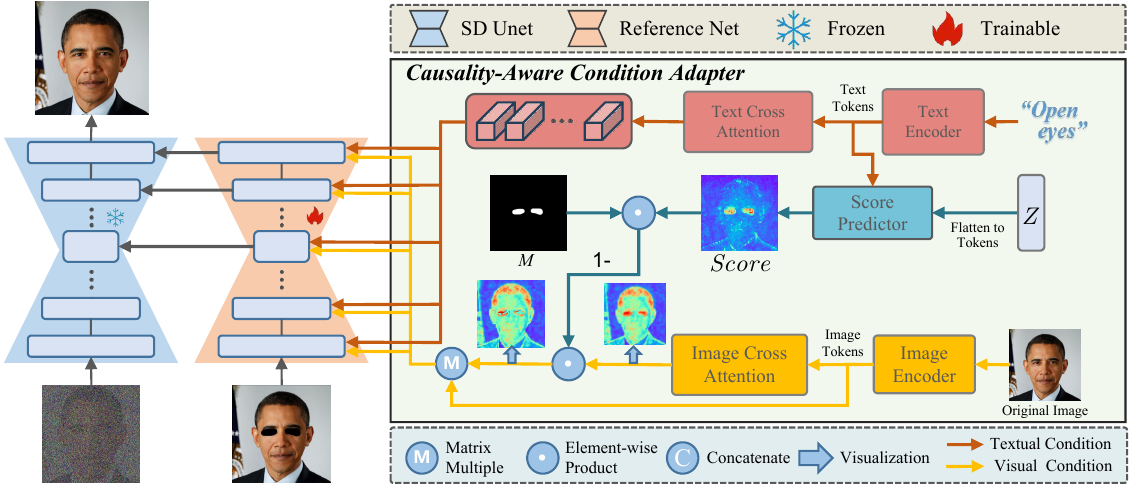}
  \caption{The training process of our method. The CA${^2}$ in the Reference Net to inject specific skin details from the original image as image embedding via an additional attention mechanism. Furthermore, the CA${^2}$ employs an adaptive score map that dynamically modulates the intensity of the visual condition, preventing conflict the causality modeling.}
  \label{fig:overview}
\end{figure*}

Specifically, we introduce our LAMask-Caption, a dataset consisting the triples of detailed textual captions of local facial attributes, high-resolution images and attribute masks.
The overview of our LAMask-Caption construction pipeline is shown in Fig. \ref{fig:data}.
Via this framework, we collect a high-quality facial image dataset comprising 200,000 high-quality images by combining filtered images from FaceCaption-15M with selections from FFHQ and CelebMask-HQ datasets.

We employ Multimodal Large Language Models (MLLMs) \cite{chen2023sharegpt4v, li2024mgm} to generate local textual captions, encouraging diverse responses that describe the face images from various perspectives, including direct, indirect, and subjective perceptions. 
Additionally, we use a fine-tuned BiSeNet \cite{yu2018bisenet} to create segmentation masks for 19 facial attributes. 
Hereto, caption-mask pairs corresponding to local facial regions could be acquired, forming the core component of the proposed LAMask-Caption. 

\subsection{Causality-Aware Condition Adapter ($\text{CA}^{2}$)}
One naive approach for injecting skin detail as a visual condition into a diffusion model is usually achieved through cross-attention, which requires parallel addition of cross-attention modules for the original image embedding, akin to IP-Adapter \cite{ye2023ip, wang2024instantid}.
However, we argue that the direct injection of visual cross-attention would lead to over-reliance on the visual condition during training while ignoring textual editing conditions \cite{jeong2024visual, qi2024deadiff}. 
To this end, we propose the novel Causality-Aware Condition Adapter (CA$^{2}$), as shown in Fig. \ref{fig:overview}, which injects specific skin details from the original image as image embedding through an additional attention mechanism, and adaptively adjusts the intensity of visual condition injection.
The adjustment is conducted based on the influence of the textual prompt on the existing features, aiming to balance the impact of textual and visual conditions.
The adapter encodes the contextual causality between the main editing region and specific skin details, while preventing visual-textual condition conflicts.

In our proposed $\text{CA}^{2}$, both the vision and text encoders of a pretrained CLIP are utilized for the feature extraction, formulated as:
\vspace{-3pt}
\begin{equation}
\left\{
\begin{array}{l}
     {f}_{txt} = CLIP_{txt}(txt) \in\mathbb{R}^{n_{t}\times c_{t}}\\
     f_{vis} = CLIP_{vis}(x) \in\mathbb{R}^{n_{v}\times c_{v}} \\
\end{array} 
\right.
 \label{eq:clip}
\end{equation}
where $n_{t}, n_{v}$ denote the numbers of text and visual tokens, and $c_{t}, c_{v}$ are the dimensions of text and vision tokens. $CLIP_{txt}, CLIP_{vis}$ are the CLIP text and vision encoders, respectively. $x$ is the original image.


Subsequently, we intend to use the textual pooling token $f_{txt}^{pool} \in \mathbb{R}^{1 \times c_{t}}$ along with the diffusion model's latent features $Z\in \mathbb{R}^{ n_{z} \times c_{z}}$ to predict textual importance scores. We spatially replicate $f_{txt}^{pool}$ to $f^{s}_{txt}\in\mathbb{R}^{n{z}\times c_{t}}$ to align the token numbers, where $n_{z}$ is the token number of $Z$.

To obtain the score that is used to weight the importance of visual condition, a simple two-layer MLP with a softmax activation function is constructed as the score predictor. The score takes the concatenation of textual class token and diffusion latent features along the channel dimension as input and then predicted following:
\begin{equation}
    Score = \mathcal{S}(\mathrm{Concat}(Z, f^{s}_{txt})) \\
\label{eq:score}
\end{equation}
where $\mathcal{S}(\cdot)$ is the score predictor, $Score \in\mathbb{R}^{n_{z}}$ and then it will be reshaped to match the spatial dimension of the latent feature.
Meanwhile, the visual cross-attention map is calculated as:
\vspace{-6pt}
\begin{equation}
        A_{vis} = \mathrm{Softmax}(\frac{\mathbf{Q}(\mathbf{K}_{vis})^\top}{\sqrt{d}})
\label{feature}
\end{equation}
where $\mathbf{Q} = Z \phantom{\cdot} W^{Q}$, $\mathbf{K}_{vis} = f_{vis} W^{K}_{vis}$, are the query of latent feature $Z$ and the key from vision feature $f_{vis}$, respectively. $W^{Q}$ and $W^{K}_{vis}$ are the corresponding weight matrices. The query matrix of the vision feature is the same as that of text cross-attention. 
Pixels with higher textual importance scores should have their vision attention suppressed, as this indicates stronger textual editing. Conversely, pixels with lower scores should receive higher vision attention to enhance dependence on the original image.
Therefore, we intend to suppress the vision attention values within the mask region according to the obtained $\mathrm{Score}$ as:

\begin{equation}
\begin{aligned}
    A^s_{vis} &= A_{vis} \odot (1- Score\odot M)  \\ 
    F_{vis}^{s} &= A^s_{vis} \mathbf{V}_{vis}
\end{aligned}
\label{eq:As}
\end{equation}


where $\odot$ denotes the Element-wise product, $M$ is the input mask that has been downsampled to the same spatial resolution as the $Score$ prior to the flattened representation. $\mathbf{V}_{vis} =  f_{vis} W^{V}_{vis}$ denotes the value of vision feature in cross-attention. 
Eventually, the latent feature processed by our CA$^2$ can be computed as:


\begin{equation}
\begin{aligned}
    F_{txt} &= \mathrm{Softmax}(\frac{\mathbf{Q}(\mathbf{K}_{txt})^\top}{\sqrt{d}})\mathbf{V}_{txt} \\[-6pt]
    Z_{s} &= F_{txt} + F^{s}_{vis}
\end{aligned}
\label{eq:As}
\end{equation}

where $\mathbf{K}_{txt}$ and $\mathbf{V}_{txt}$ denote the key and value of $f_{txt}$ in Eq. (\ref{eq:clip}), respectively.


\subsection{Skin Transition Frequency Guidance (STFG)}

While CA$^2$ preserves skin details in the main editing areas, real-world facial editing often uses imprecise masks, leading to unnatural transitions in `boundary regions'. These smooth skin areas are sensitive to low-frequency changes. To address this, we introduce a sampling guidance technique for low-frequency components during denoising, to produce natural transitions in these regions.

Specifically, given the localization and semantic representation capabilities of textual cross-attention maps in diffusion models to identify `boundary regions'. The mean of attention maps, i.e., $\overline{A}_{txt}$ is computed across all text tokens and attention layers. 
We identify the `boundary regions' as regions within the mask $M$ where the attention values on $\overline{A}_{txt}$ are below a threshold $\gamma (\overline{A}_{txt}, M)$. The indexes $Idx$ of the pixels belonging to `boundary region' is represented as:
\begin{equation}
\begin{array}{c} 
Idx = \{ (i, j) | \overline{A}_{txt} (i, j)  \le \gamma (\overline{A}_{txt}, M) \} \\
\gamma (\overline{A}_{txt}, M)=\mu (\overline{A}_{txt}\circ M)-\sigma(\overline{A}_{txt}\circ M) \\
\end{array}
\label{eq:idx}
\end{equation}
where $\overline{A}_{txt}\circ M$ represents the elements of $\overline{A}_{txt}$ within the mask $M$, $\mu(\cdot)$ and $\sigma(\cdot)$ denote the operators of mean and standard deviation.

\begin{figure*}[t]
  \centering
  \captionsetup{skip=5pt}
  \includegraphics[width=1.0\linewidth]{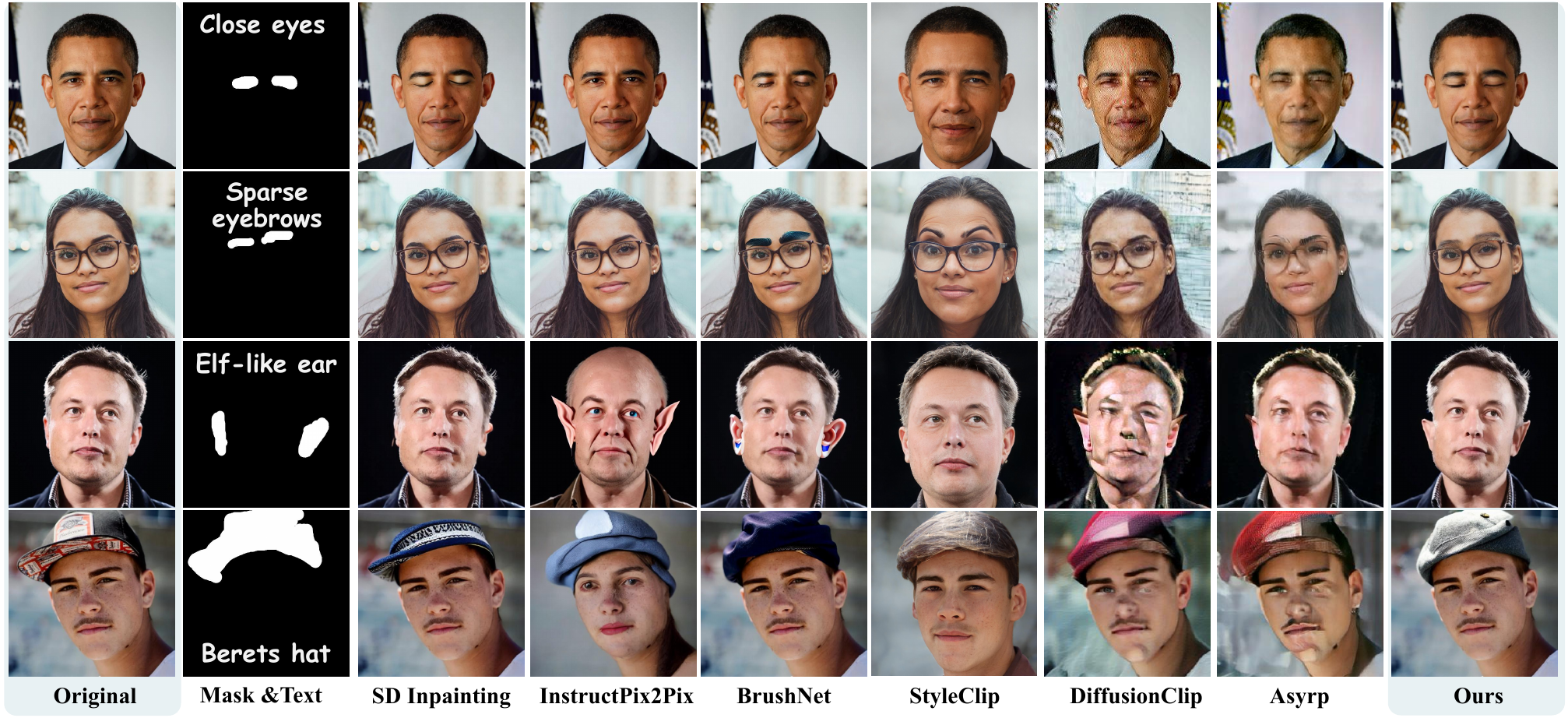}
  \caption{Qualitative comparison on local facial attributes editing. Compared with zero-shot methods (i.e. SD inpainting \cite{wang2022high}, InstructPix2Pix \cite{brooks2023instructpix2pix}, BrushNet \cite{ju2024brushnet}) and the facial editing methods ( StyleClip \cite{patashnik2021styleclip}, Diffusionclip \cite{kim2022diffusionclip}, Asyrp \cite{KwonJU23} ), our approach not only aligns the edited parts with the text prompts, but also better preserves the information from the original image.}
  \label{fig:sota}
\end{figure*}

We further employ frequency guidance in the Fourier domain to selectively enhance low-frequency similarity on the estimated latent, i.e., designing a guidance function to pixel-wisely align the low-frequencies between the original noisy latent $z_{t}$ and the predicted latent $\widehat{z}_{t}$ on each timestep $t$. Since the frequency components should be calculated on the clean latent, we estimate the one-step prediction $\widehat{z}_{t\rightarrow 0}$ from $\widehat{z}_{t}$ as:
\begin{equation}
\begin{array}{c}
        \widehat{z}_{t\rightarrow 0}=\frac{\widehat{z}_{t}}{\sqrt{\bar{\alpha}_t}}-\frac{\sqrt{1-\bar{\alpha}_t}\epsilon_\theta(\widehat{z}_{t},t)}{\sqrt{\bar{\alpha}_t}}
\end{array}
\end{equation}
where $\bar{\alpha}_t$ is the hyperparameter of noise schedule parameter. Subsequently, we only keep the low-frequency components ($\frac{H}{2}<h<\frac{3H}{4} \text{and} \frac{W}{2}<w<\frac{3W}{4}$ in FFT shifted image) of $\widehat{z}_{t\rightarrow 0}$ and $z_{0}$ in the frequency domain to obtain $\widehat{z}^{\prime}_{t\rightarrow 0}$ and $z^{\prime}_{0}$, respectively. Consequently, the guidance function used to align these two can be defined as follows:
\begin{equation}
    g(z^{\prime}_{0}, \widehat{z}^{\prime}_{t\rightarrow 0}) = \frac{1}{|Idx|} \sum\limits_{(i,j)\in Idx}\left\|\widehat{z}^{'}_{t\rightarrow 0}(i,j)-z_{0}^{'}(i,j)\right\|_{2}^{2}
\end{equation}
where $|Idx|$ is the cardinality of the set $Idx$. We follow the score-based guidance \cite{song2020score}, and use $g(z^{\prime}_{0}, \widehat{z}^{\prime}_{t\rightarrow 0})$ to steer the diffusion process.
Eventually, we can update the direction of $\hat{\epsilon}_{t}$ as follows:
\begin{equation}
\begin{array}{c} 
        \hat{\epsilon}_t=\epsilon_\theta(z_t,t,txt,x)-\lambda\rho_t\nabla_{z_t}g(z^{\prime}_{0}, \widehat{z}^{\prime}_{t\rightarrow 0})
\end{array}
\label{eq:idx}
\end{equation}
where $\lambda$ is a hyperparameter of the guidance strength and $\rho_t$ denotes the noise schedule parameter of timestep $t$.

\begin{figure}[htb]
  \centering
  \captionsetup{skip=2pt}
  \includegraphics[width=1.0\linewidth]{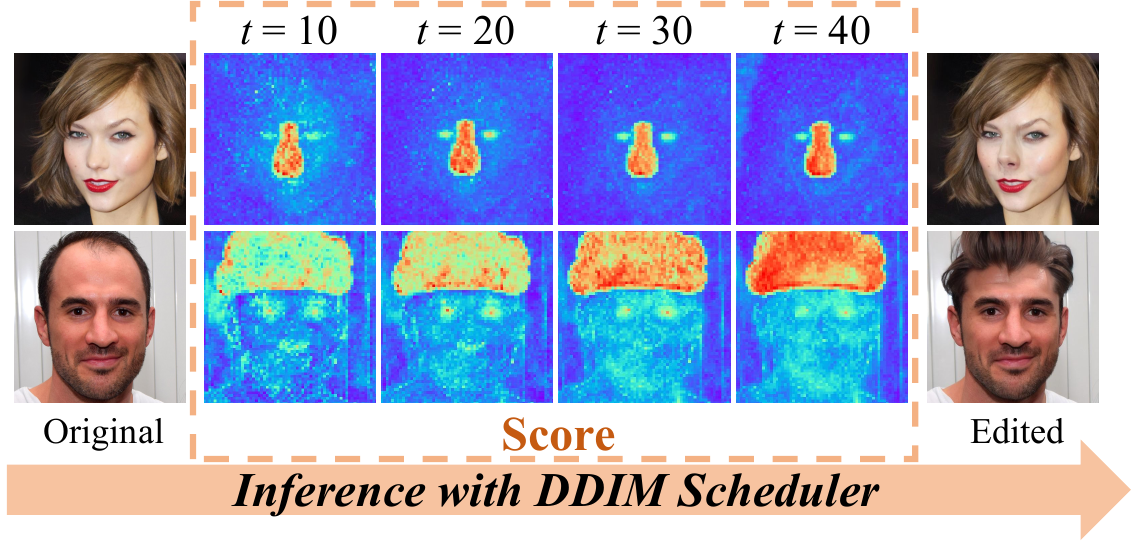} 
  \caption{The visualization of the score in CA$^{2}$ during inference. The lighter regions indicate the higher values in the maps. The DDIM scheduler with $t=50$ timesteps is used.}
  \label{fig:score}
\end{figure}

\section{Experiment}

\subsection{Evaluation Metric}

\textbf{Objective Metrics.} To comprehensively evaluate the performance of different methods on the task of local facial attributes editing, we utilize \textbf{FID} / \textbf{Local-FID} \cite{heusel2017gans}, \textbf{LPIPS} \cite{zhang2018unreasonable}, \textbf{identity similarity (ID)}, \textbf{MPS} \cite{Zhang_2024_CVPR} and \textbf{HPSv2} \cite{wu2023human} as evaluation metrics.
\textbf{FID} and \textbf{LPIPS} are used to provide an estimate of image fidelity. It's important to note that in this specific task, unlike general image generation, lower LPIPS values indicate higher fidelity. \textbf{MPS} and \textbf{HPSv2} are more effective and comprehensive zero-shot objective evaluation metrics on text-image alignment and human aesthetics preferences. ID evaluates the face identity between the results and the original images.

\textbf{User Study.} 
Besides comparisons on objective metrics, we also conduct a user study via pairwise comparisons to determine whether our method is preferred by humans. 
The generation results are evaluated on three dimensions: face fidelity (FF), text-attribute consistency (TAC), and human preference (HP).

\subsection{Experimental Setup}
\label{sec:expriment}
\textbf{Benchmark.} 
As this work serves as one of the text-guided local facial attribute editing, we introduce \textit{FFLEBench}, i.e., one pioneering benchmark evaluation dataset for this task, motivated by the lack of corresponding benchmark and evaluation dataset. \textit{FFLEBench} comprises a total of 15,000 samples drawn from FFHQ, accompanied by the local masks and the corresponding textual captions.
Note that the samples drawn from FFHQ to construct the \textit{FFLEBench} are independent with those used for training.
The masks are the convex hull or the dilation of the segmentation masks, aiming to imitate the rough mask input.

\textbf{Implementation Details.} 
All the cross-attention maps and the score map are upsampled to the resolution of 64 $\times$ 64. To preserve the original information in the regions outside the mask, we blend the latent variable following Blended Diffusion \cite{avrahami2022blended}. 

\begin{table*}[htbp]
\centering
\caption{Quantitative comparisons between the state-of-the-art methods and ours. "Ours vs." indicates the proportion of users who prefer our proposed method over a comparative approach.
The proportion in user study exceeding 50\% indicates that our method outperforms the counterpart.
MPS exceeding 1.00 indicates that our method outperforms the counterpart. local-FID (L-FID) is computed within the bounding box of the mask region. Number in ``( )" is the time required for single attribute fine-tuning of facial editing methods.
}
\label{tab:editing_performance}
\scalebox{0.95}{
\begin{tabular}{lcccccccc}
\midrule
\multirow{2}{*}{\textbf{Method}} & \multicolumn{5}{c}{\textbf{Objective Metrics}} & \multicolumn{3}{c}{\textbf{User study (Ours vs. )}}\\
\cmidrule(r){2-6}\cmidrule(r){7-9}
& FID/L-FID ($\downarrow$) & LPIPS ($\downarrow$) &HPSv2($\uparrow$) &ID ($\uparrow$)   & MPS ($\uparrow$)   & FF ($\uparrow$) & TAC ($\uparrow$) &HP ($\uparrow$)\\ 
\midrule
SD Inpainting         & \textbf{3.11/1.61} & 0.175& 0.248 & 0.63       & 1.03     &86.05\% &79.32\% &77.88\%\\
BrushNet              & 5.45/2.30          & 0.285& 0.254 & 0.59       & 1.34     &86.05\% &83.17\% &82.69\%\\
IntructPix2Pix        & 8.36/5.36         &\underline{0.160} & 0.263 & 0.67       & 1.03     &87.98\% &83.65\% &85.09\%\\
\midrule
DiffusionClip (310s)  &8.19/5.68           & 0.301&0.257 &\textbf{0.73}&1.13       &93.56\% &68.29\%        &92.31 \%\\
Asyrp (408s)          &8.11/6.32           & 0.260&0.240 & 0.62        &1.80       &86.05\% &63.29\%        &84.28 \% \\
StyleClip (40s)       &6.38/4.83           & 0.249&\underline{0.263} & 0.63        &1.09       &93.68\% &83.17\%        &68.38\%\\
\midrule
\rowcolor{gray!20} \textbf{Ours}  & \underline{4.81/1.99}  & \textbf{0.085} & \textbf{0.264} & \underline{0.72} & /    & / & / & / \\
\midrule
\end{tabular}
}
\end{table*}


\begin{figure}[htb]
  \centering
  \captionsetup{skip=0pt}
  \includegraphics[width=1.0\linewidth]{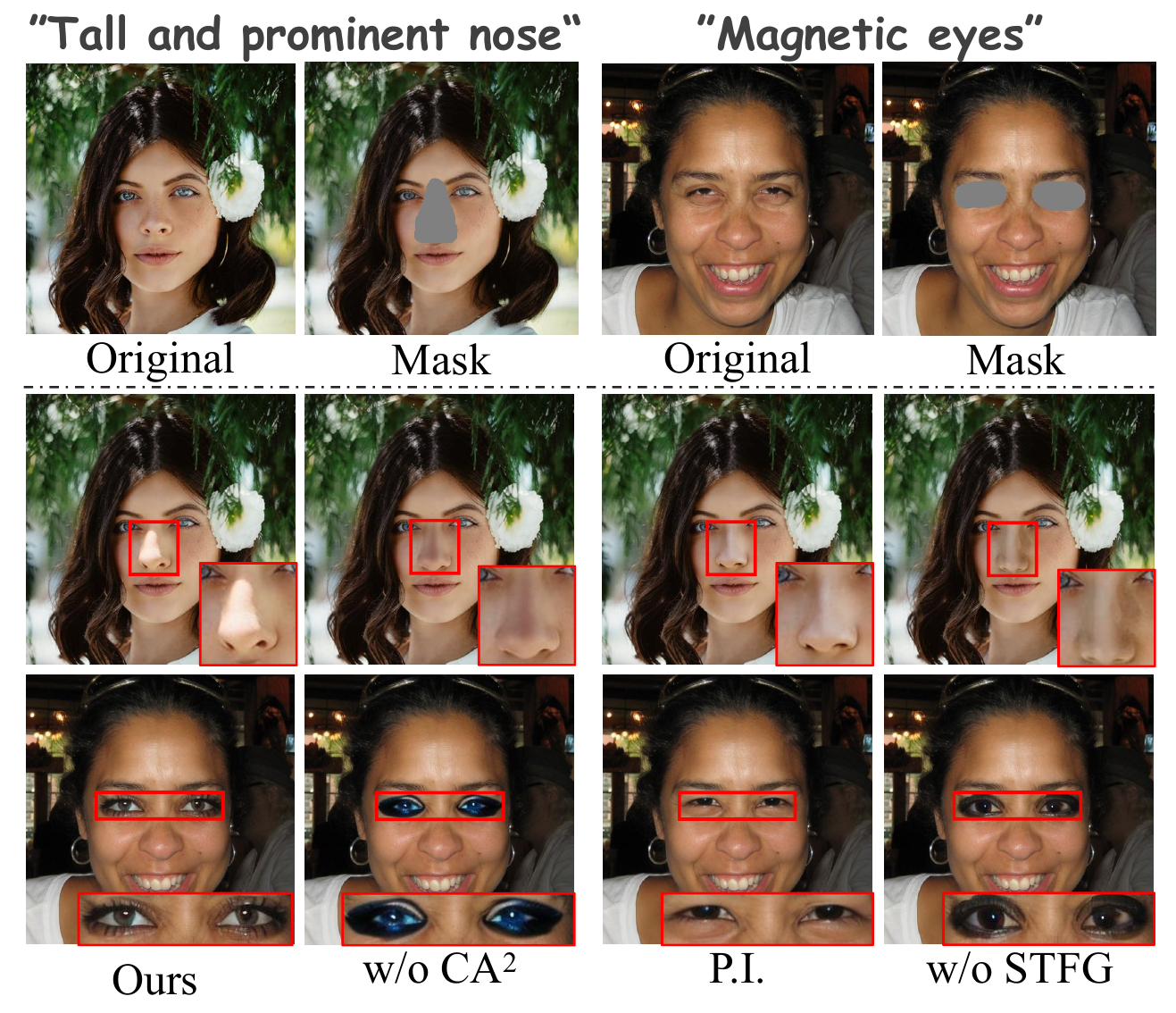} 
  \caption{Ablation study of our modules. ``Parallel Injection (P.I.)" removes the score in Eq. (\ref{eq:score}), “w/o CA${^2}$” removes the CA${^2}$ but preserves textual condition.}
  \label{fig:abla}
\end{figure}

\subsection{Quantitative Experiment Results}
We quantitatively evaluate our method on FLEBench, compared with baseline models using both objective metrics and user study.
As shown in Tab. \ref{tab:editing_performance}, the proposed method surpasses the compared methods except for the FID of Stable Diffusion Inpainting. Particularly, better performance on FID and LPIPS indicates that our method can edit facial attributes with higher fidelity.
Due to its tendency to neglect the text prompt to maintain high fidelity, SD inpainting exhibits the lowest FID score. 
Our approach outperforms on the MPS and HPS v2 metrics, indicating our edits align with human aesthetics and maintain textual consistency. 
All the observation highlights the strength of our approach in preserving visual coherence and effectively capturing the textual guidance during the editing.
In addition, our approach achieves better local attribute editing results without requiring the extra fine-tuning time for different attributes, which is needed by other facial editing methods (fine-tuning time shown in brackets after the method names).

In the user study, the percentages represent the proportion of users who prefer our method over others. As shown in Tab. \ref{tab:editing_performance}, our method attains the top rank compared to the other inpainting and face editing methods.

\begin{figure}[htb]
  \centering
  \captionsetup{skip=2pt}
  \includegraphics[width=1.0\linewidth]{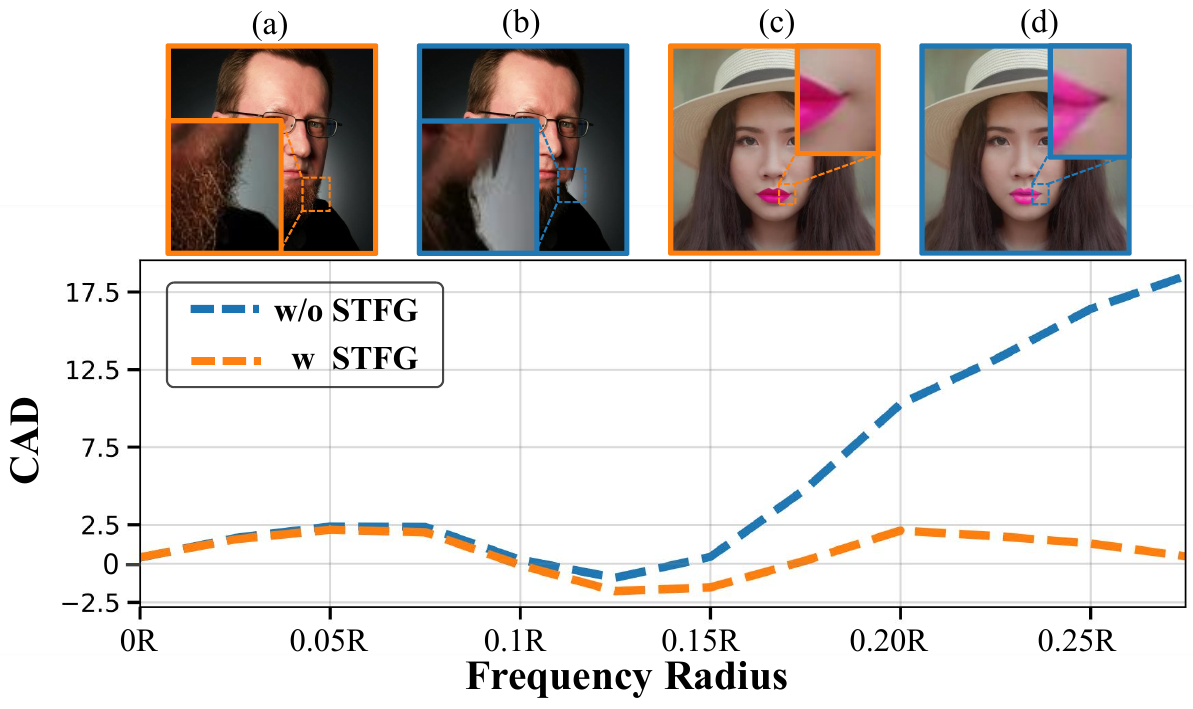} 
  \caption{Cumulative amplitude difference (CAD) in the Fourier domain between the sampled and the original images is calculated within the mask region, 
  specific to the FFT-shifted image with a radius representing by the $x$-axis (R is the max of Frequency Radius).
  (a) and (c) are the sampled images with (`w') STFG. (b) and (d) are the sampled images without (`w/o') STFG.}
  \label{fig:STFG}
\end{figure}

\subsection{Qualitative Experiment Results}

\textbf{Comparison with the SOTAs.} 
In Fig.\ref{fig:sota}, our method is qualitatively compared with the state-of-the-art (SOTA) methods across the local facial attributes, such as eyes, ears, and accessories. Other manipulation results are in the supplementary materials.
Prompt neglect is an issue for other methods that sometimes struggle to modify local attributes according to textual descriptions, as evident in the ``sparse eyebrows" example (second row).
While they can capture text semantics in some cases, they miss the original images' specific skin details, compromising overall fidelity.

In addition, InstructPix2Pix, and the facial editing methods (i.e. StyleClip, DiffusionClip and Asyrp) exhibit undesirable content leakage into adjacent regions, resulting in effects beyond the intended target area.
In contrast to prior limitations, our method enhances consistency between edited regions and text prompts, while preserving original skin details by understanding the contextual causality between generation and source image information.

\textbf{Analysis of the Score in CA${^2}$.}
We visualize the score in Eq.(\ref{eq:score}) during inference to explore how our CA$^2$ dynamically prevents the conflict between visual and textual condition. 
The lighter region of a score map corresponds to higher values, which in turn indicates less injection of image features in those regions. 
As shown in Fig. \ref{fig:score}, our model initially exhibits more attention to the image prompts in the early timesteps, i.e., it refers to the original image to maintain the skin tones. 
As the inference continues, the model relies less on the original image and generates the contents according to text.
Furthermore, it shows that the score map exhibits lower values at the edges and in regions with minimal editing, suggesting that these regions rely more heavily on the original image.
This is consistent with the motivation of the score in CA$^{2}$, which enables the model to spatially control the sensitivity of image prompts. 

\textbf{Analysis of the frequency guidance of STFG.}

To study the capacity of STFG in enhancing the low-frequency similarity during sampling, we calculate the cumulative amplitude difference in the Fourier domain between the sampled and the original images, varying as the frequency radius within the mask region.
The amplitude difference specific to our STFG (the orange line) fluctuates around zero, indicating that STFG can effectively promote low-frequency similarity between the sampled and original images, which is helpful for the skin detail preservation.
The images shown above the graph demonstrate that the artifacts in the edge region have been effectively eliminated by STFG.

\subsection{Ablation Study}
We demonstrate the effectiveness of our module through the generation qualitative quality and quantitative metrics (appendix).
As shown in the second column in Fig. \ref{fig:abla}, after removing CA$^{2}$, the variant simply follows the text instructions, making the generated content inconsistent with skin tone of original image.
The model without the score in Eq. (\ref{eq:score}) exhibited obvious content leakage and was unable to faithfully follow the text description.
It demonstrates that the score in Eq. (\ref{eq:score}) plays a role in encouraging the model to prioritize textual editing.
For the inference without our STFG, it shows that there are obvious artifacts present in the regions around the attributes, i.e., the boundary regions.

\section{Conclusion}
This paper introduces a novel inpainting technique for local facial attribute editing that overcomes the long-lasting issues in current models, i.e. the hardness of following the local facial attribute description and the lack of contextual causality modeling on mask regions.
We present a new data strategy and a Causality-Aware Condition Adapter to effectively incorporate original image skin details for causality modeling while preventing conflict between visual and textual condition. Moreover, a Skin Transition Frequency Guidance is introduced to improve the coherence of generated content around the boundaries. Extensive experiments show the superior performance of our method over current SOTA ones.

\section*{Acknowledgment}

The work was supported by the National Natural Science Foundation of China under grants no. 62276170, 82261138629, the Guangdong Basic and Applied Basic Research Foundation under grants no. 2023A1515011549, 2023A1515010688, the Science and Technology Innovation Commission of Shenzhen under grant no. JCYJ20220531101412030, and Guangdong Provincial Key Laboratory under grant no. 2023B1212060076.


\bibliography{aaai25}

\clearpage
\appendix
\begin{center}{\bf \Large Appendix}\end{center}\vspace{1em}

\section{Details of Our Proposed LAMask-caption}

To align the generated content and textual description, existing inpainting methods \cite{manukyan2023hd, mao2023magedit} reweigh the attention scores during the inference stage.
They hold the key insight that adjusting the noise latent feature to attain higher cross-attention values to enhance its alignment with the specific text prompt.
However, their approach is based on that the base model already has prior knowledge about the target editing content. The major reason that current diffusion models fail to generalize to local face inpainting is the lack of precise textual captions to the images, as mainstream diffusion models are mainly trained on large-scale natural image datasets such as Laion-2B \cite{schuhmann2022laion}, Laion-Aesthetics\cite{lin2014microsoft}.
While existing CelebA-dialog \cite{jiang2021talkedit}, FaceCaption-15M \cite{dai202415m} and FLIP-80M \cite{li2024flip} mainly focus on overall attributes (i.e. age, skin) rather than local facial attributes.

\begin{figure*}[!ht]
  \centering
  \includegraphics[width=1.0\linewidth]{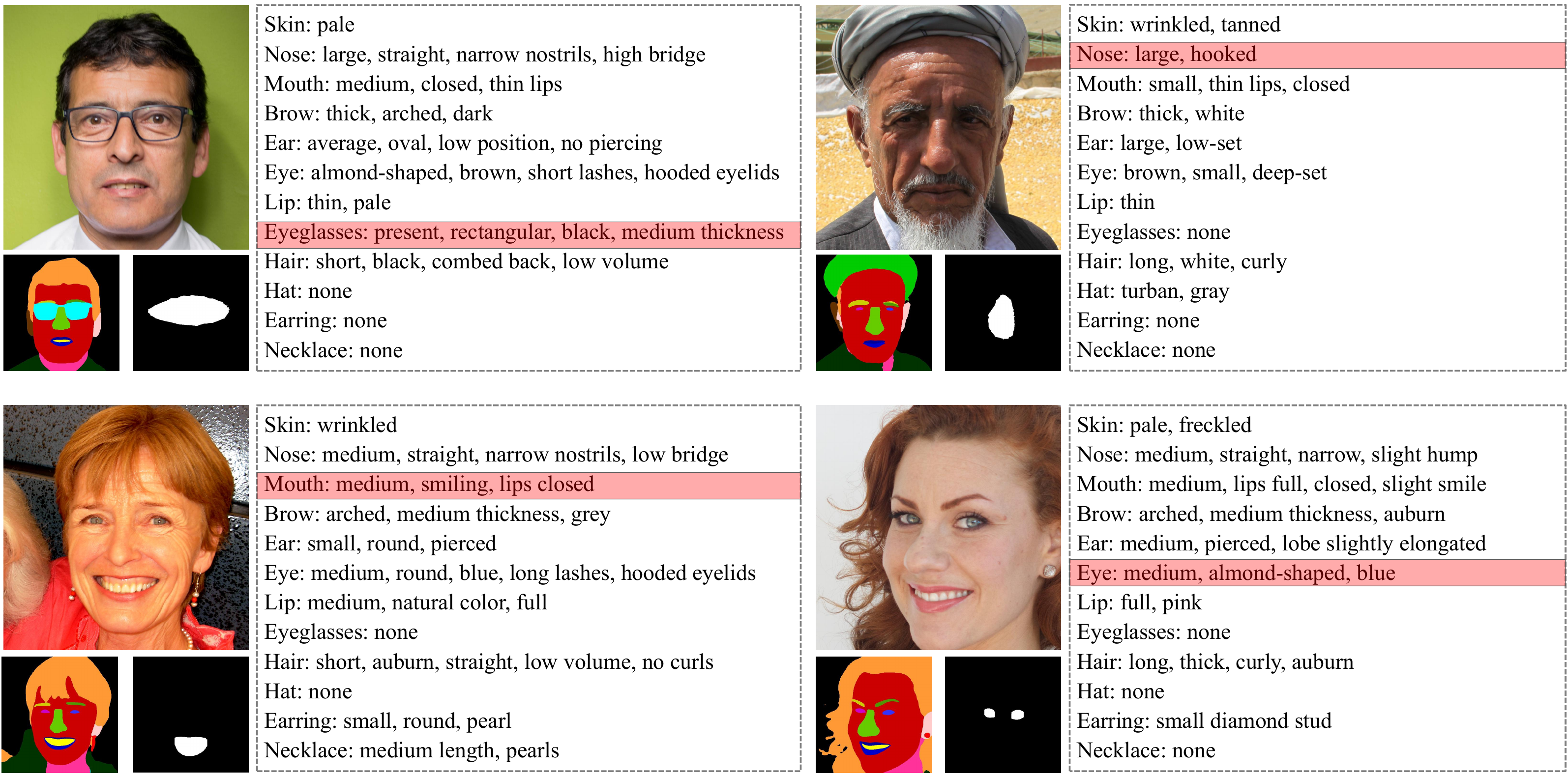}
  \caption{Examples of our proposed LAMask-Caption.}
  \label{fig:bench}
\end{figure*}

Therefore, we proposed LAMask-Caption that mainly consists of face images, textual descriptions of local facial attributes and the corresponding segmentation mask of the regions.
The examples of our LAMask-Caption are shown in Fig. \ref{fig:bench}.

\textbf{Inpainting Masks of Local Facial Attributes.}
Firstly, segmentation masks for facial attributes are generated to fit the training paradigm inpainting. 
Specifically, we used a fine-tuned BiSeNet \cite{yu2018bisenet} to segment a face into 19 parts (as shown in Fig. \ref{fig:bench}), where each local region mask would have a corresponding caption generated by the MLLMs. 
Given that the precise masks may lead the model to learn trivial solutions during training \cite{yang2023paint}, resulting in artifacts around the boundary of the masked region in the inpainting content.
Therefore, we use the image erosion algorithm and Bessel curve fitting to the bounding box of the mask as pre-processing methods to generate mask augmentation.

\textbf{Caption of local facial attributes.} 
To obtain specific local textual captions, the Multimodal Large Language Models (MLLMs) ShareGPT-4V \cite{chen2023sharegpt4v}, MGM \cite{li2024mgm} are employed for caption generation. The MLLMs is given a textual prompt and a face image, and requested to generate captions for each corresponding local region. To enhance textual diversity, the MLLM is encouraged to generate the responses encompassing various perspectives including direct appearance descriptions, indirect appearance descriptions (e.g. elf-like ears), and subjective perceptual feelings (e.g. complex eyes as if he/she holds secrets that cannot be unravelled).

\begin{figure}[!ht]
  \centering
  \includegraphics[width=1.0\linewidth]{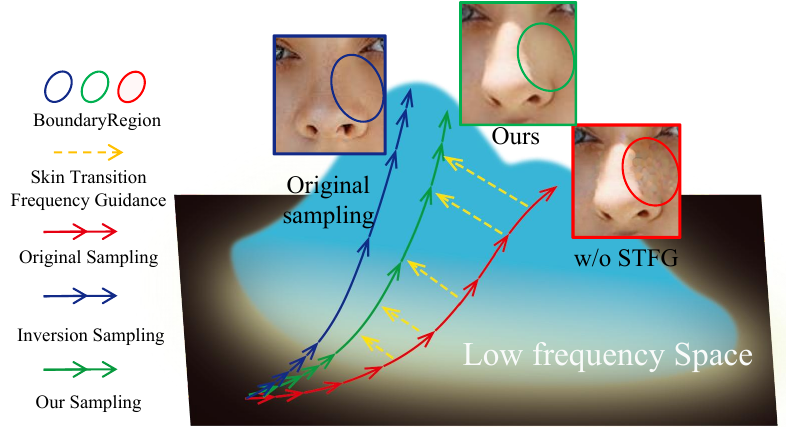}
  \caption{The illustration of how STFG works. During the sampling process, through the guidance of STFG, the pixels in the ``boundary region" of the latent  gradually approach to that of original image in the low-frequency domain.}
  \label{fig:STFG_inference}
\end{figure}


\section{Details of the Skin Transition Frequency Guidance (STFG) }

We provide more details of our STFG in this section. As shown in Fig. \ref{fig:STFG_inference}, our proposed STFG can reduce artifacts and produce natural transitions in the ``boundary regions".

In the main body, we propose to leverage textual cross-attention maps of diffusion models as a prior to localize the ``boundary regions", as the attention maps exhibit outstanding localization performances and semantic understanding ability \cite{simsar2023lime}.
Specifically, for each text token $j = {1, \cdots, l}$, where $l$ is the number of text tokens, we first upsample each textual cross-attention map $A_{txt}^{i}[j]$ in the Reference Net to the size $\mathbf{H} \times \mathbf{W}$, and compute their mean as: 
\begin{equation}
\begin{split}
        \overline{A}_{txt} = \frac{1}{m \cdot l} \sum_{i=1}^{m}{ \sum^{l}_{j=1} {(A_{txt}^{i}[j])} }
\end{split}
\label{feature}
\end{equation}
where $m$ denotes the number of textual cross-attention layers of the Reference Net and $A_{txt}^{i}[j]$ represents the attention map of the $j$-th textual token from the $i$-th layer.
To separate the major editing regions and the transition regions within the coarse mask, we define the regions within the mask that are minimally influenced by the text prompt as the ``boundary regions".
We propose to identify the indexes $Idx$ of all the elements belonging to ``boundary region" according to:
\begin{equation}
\begin{array}{c} 
        Idx = \{ (i, j) | \overline{A}_{txt} (i, j)  \le \mu - \sigma \}
\end{array}
\label{eq:idx}
\end{equation}
where $\mu$ and $\sigma$ denote the mean and the variance of $\overline{A}_{txt}$:
\begin{equation}
\begin{array}{c} 
   \mu = \frac{1}{\mathbf{H}\cdot \mathbf{W}}\sum_{i,j}  {\overline{A}_{txt}(i,j) \odot M},\quad  \\
   \sigma = \sqrt{\frac{{\sum_{i,j}{(\overline{A}_{txt}(i, j) \odot M}-\mu)^2}}{\mathbf{H} \cdot \mathbf{W}}} 
\end{array}
\label{eq:As}
\end{equation}

To preserve skin details, we propose to further enhance the similarity of these boundary regions with the low-frequency components of the original image. Since the frequency component should be calculated based on the clean latent, we first estimate the  $\widehat{z}_{t\rightarrow 0}$ from $\widehat{z}_{t}$ as:
\begin{equation}
\begin{array}{c} 
        \widehat{z}_{t\rightarrow 0}=\frac{\widehat{z}_{t}}{\sqrt{\bar{\alpha}_t}}-\frac{\sqrt{1-\bar{\alpha}_t}\epsilon_\theta(\widehat{z}_{t},t)}{\sqrt{\bar{\alpha}_t}}
\end{array}
\label{eq:one-step}
\end{equation}

To pixel-wisely align the low-frequencies between the original noisy latent $z_{t}$ and the predicted latent $\widehat{z}_{t}$, mathematically, we propose to keep low-frequency components of both the estimated latent $\widehat{z}_{t \rightarrow 0}$ and the original latent $z_{0}$, which are first obtained as:
\begin{equation}
    \begin{aligned}
        &\mathcal{F}(\widehat{z}_{t\rightarrow 0}) = \mathrm{FFT}(\widehat{z}_{t\rightarrow 0}), \quad
        \mathcal{F}(z_{0}) = \mathrm{FFT}(z_{0}) \\
        &\mathcal{F}^{\prime}(\widehat{z}_{t\rightarrow 0}) = \mathcal{F}(\widehat{z}_{t\rightarrow 0}) \odot \mathbf{1}_t, \quad
        \mathcal{F}^{\prime}(z_{0}) = \mathcal{F}(z_{0}) \odot \mathbf{1}_{t} \\
        &\widehat{z}^{\prime}_{t\rightarrow 0} = \mathrm{IFFT}(\mathcal{F}^{\prime}(\widehat{z}_{t\rightarrow 0})), \quad
        z^{\prime}_{0} = \mathrm{IFFT}(\mathcal{F}^{\prime}(z_{0}))
    \end{aligned}
\end{equation}


where $\mathrm{FFT}(\cdot)$ and $\mathrm{IFFT}(\cdot)$ are Fourier transform and inverse Fourier transform, respectively, $\mathbf{1}_t(i, j) = [\frac{H}{2}<i<\frac{3H}{4} \text{and} \frac{W}{2}<j<\frac{3W}{4}]$ is a Fourier mask, and designed as a characteristic function.

We then employ the guidance in the Fourier domain to selectively enhance low-frequency similarity on the estimated latent, i.e., a guidance function $g$ is designed to steer the diffusion process and defined as follows:
\begin{equation}
\begin{array}{c} 
        g(z^{\prime}_{0}, \widehat{z}^{\prime}_{t\rightarrow 0}) = \frac{1}{|Idx|} \sum\limits_{(i,j)\in Idx}\left\|\widehat{z}^{'}_{t\rightarrow 0}(i,j)-z_{0}^{'}(i,j)\right\|_{2}^{2},
\end{array}
\label{eq:gfunc}
\end{equation}
Eventually, the update direction $\hat{\epsilon}_{t}$ is defined as follows:
\begin{equation}
\begin{array}{c} 
        \hat{\epsilon}_t=\epsilon_\theta(z_t,t,txt,x)-\lambda\rho_t\nabla_{z_t}g(z^{\prime}_{0}, \widehat{z}^{\prime}_{t\rightarrow 0})
\end{array}
\label{eq:update}
\end{equation}
where $\lambda$ is a hyperparameter of the guidance strength and $\sigma_t$ denotes the noise schedule parameter of the timestep $t$.

\begin{figure*}[!ht]
  \centering
  \includegraphics[width=1.0\linewidth]{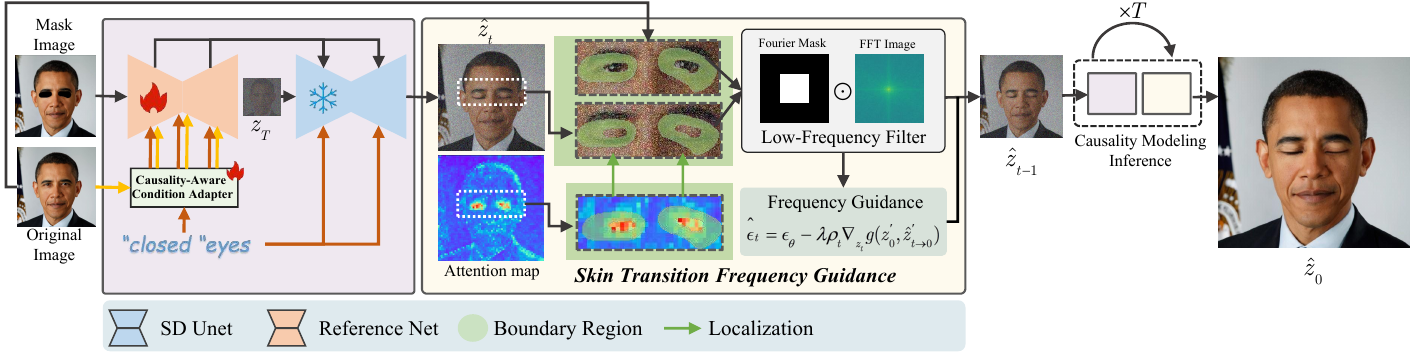}
  \caption{Our inference pipeline. }
  \label{fig:overall}
\end{figure*}

\section{Implement Details}

During training, to enable a classifier-free guidance, we follow \cite{ye2023ip} and set a probability of 0.05 to drop text or image.
We use the DDIM scheduler over $T = 50$ for denoising sampling during inference, maintaining a classifier-free guidance scale of 7.5. During inference, we utilize our STFG strategy to modify the latent variable on the ``boundary regions". For the regions out of mask, we blend the latent variable following Blended Diffusion \cite{avrahami2022blended} to preserve region features outside the mask. The whole training process and inference process are shown in Algorithms \ref{alg2} and \ref{alg1} respectively. In addition, we visualize the how to combine CA$^2$ and STFG to obtain the final inpainting result during inference in Fig. \ref{fig:overall}.

\begin{algorithm}[h!]
\renewcommand{\algorithmicrequire}{\textbf{Input:}}
\renewcommand{\algorithmicensure}{\textbf{Output:}}
\caption{Training with our CA-Edit}
\label{alg2} 
\begin{algorithmic}[1]
    \REPEAT
    \STATE Take \{latent variable $z_{0}$, vision condition $c_{vis}$, text condition $c_{txt}$ and Mask $M$\} from LAMask-Caption dataset.
    \STATE Obtain $z_{0} \sim q(z_{0}) $, $t \sim \text{Uniform}(\{1,\dots, T\})$, $\epsilon\sim\mathcal{N}(\mathbf{0},\mathbf{I})$.
    \STATE Obtain the visual and textual conditions $f_{vis}$, $f_{txt}$ in Eq.(2) in the main body.
    \STATE Use the $Score$ in Eq.(3) to weigh the importance of the visual condition spatially in Eq.(5).
    \STATE Take gradient descent step based on the loss $\mathcal{L}$ in Eq.(1) in the main body, where $c$ is replaced with $\{f_{vis}, f_{txt}, M\}$.
    \UNTIL converged
\end{algorithmic} 
\end{algorithm}

When our algorithm is compared with the mask-free methods that require the textual prompts of both original and target images, ``face" and ``face with ..." are provided as the original and target textual prompts. As for InstructPix2Pix \cite{brooks2023instructpix2pix}, i.e., an instruction-based image editing method, we utilize editing instructions such as “make” and “change” to manipulate images.

\section{Details of our proposed Benchmark \textit{FFLEBench}}

Current datasets for text-based image editing methods primarily exclude local attributes of a face.
To enable a diffusion model to generalize well to the text-driven local facial attributes editing, we 
construct the dataset, i.e., LAMask-Caption for this specific task.
For a more detailed evaluation of our method, we follow the construction pipeline to develop a benchmark dataset \textit{FFLEBench}, consisting of 15,000 images sourced from FFHQ \cite{karras2019style}. 
For masks of a human face and its various parts, including masks of skin, eyes, nose, hair, etc., these masks are processed through data augmentation (i.e. convex hull or dilation) to imitate the rough masks in the real-world application. For the text descriptions of the corresponding attributes, it encompasses direct appearance descriptions, indirect appearance descriptions and subjective perceptual feelings.

\begin{algorithm}[h!]
\renewcommand{\algorithmicrequire}{\textbf{Input:}}
\renewcommand{\algorithmicensure}{\textbf{Output:}}
\caption{Inference with our CA-Edit}
\label{alg1} 
\begin{algorithmic}[1]
    \REQUIRE Diffusion steps $T$, noisy latent $z_{T}$, original latent $z_{0}$, target text description $txt$, input mask $M$, our trained Inpainting models with parameter $\theta$, text prompt and image prompt $f_{txt}, f_{img}$.
    \ENSURE Final edited latent $z_{0}$
    \FOR{$t = T$ to 1}
    \STATE Perform $ \widehat{z}_{t-1} = \widehat{z}_{t} - \epsilon_{\theta}(t, f_{txt}, f_{img},\phantom{\cdot} $M$, z_{t})$, collect the textual attention maps $\{A^{i}_{txt}, i = 1,\cdots, token\_length\}$ during the denoising process.
    \STATE Obtain the indexes $Idx$ of the ``boundary region" on $\widehat{z}_{t-1}$ using Eq.\ref{eq:idx} and Eq.\ref{eq:As}.
    \STATE Obtain the one-step prediction $ \widehat{z}_{t\rightarrow 0}$ of $ \widehat{z}_{t}$ using Eq. \ref{eq:one-step}.
    \STATE Use the guidance function $g$ in Eq.\ref{eq:gfunc} to measure the low-frequency similarity between $ \widehat{z}^{'}_{t-1}$ and $ {z}^{'}_{0}$.
    \STATE Update the noise latent of $\epsilon_{\theta}$ with function $g$ (Eq.\ref{eq:update}).
    \ENDFOR
\end{algorithmic} 
\end{algorithm}

\section{Extended Qualitative Results}

\begin{figure*}[!ht]
  \centering
  \includegraphics[width=1.0\linewidth]{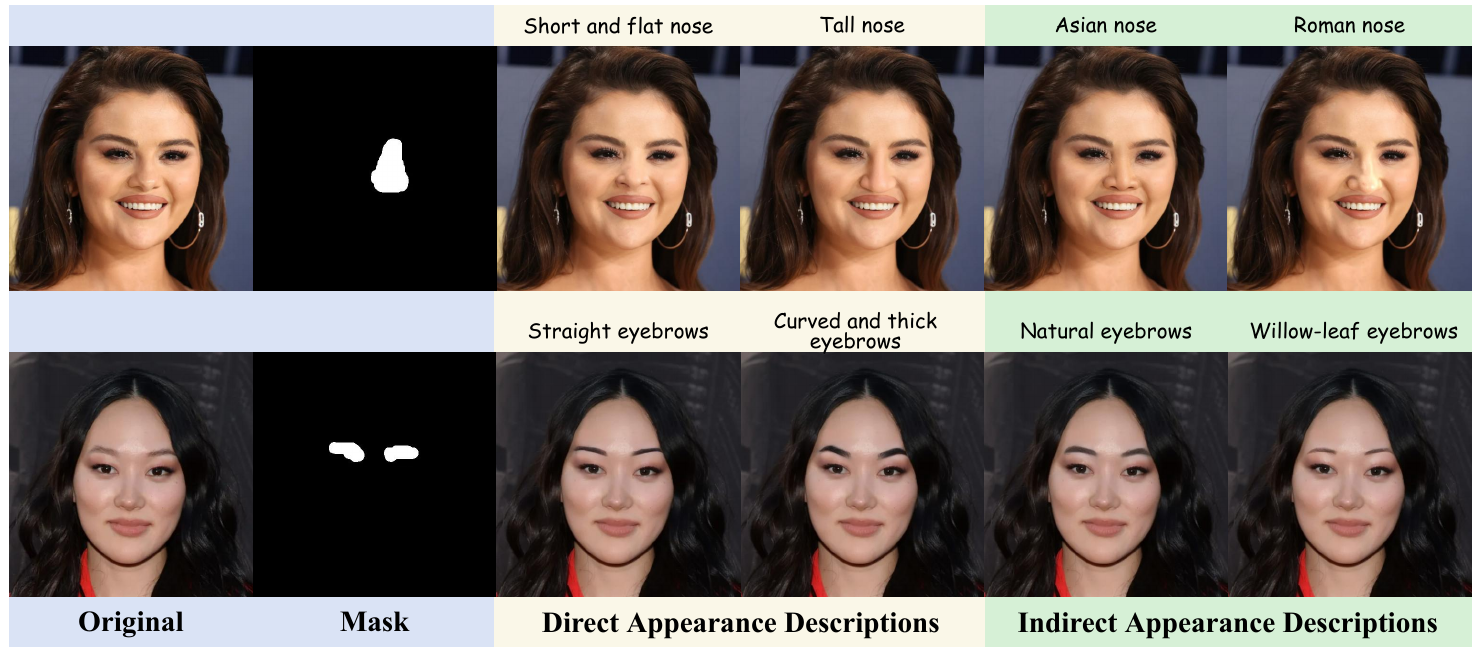}
  \caption{The inpainting results under diverse textual descriptions. Our method can faithfully handle intricate texts in different scenarios, including both direct descriptions and indirect descriptions.}
  \label{fig:multi_txt}
\end{figure*}

\textbf{Results with Diverse Description.} 
To showcase the capability of our proposed approach in following intricate instructions, we present the generated images under the same input mask for various text descriptions, including both direct and indirect ones. As shown in Fig. \ref{fig:multi_txt}, the output images highlight the adaptability of our method in accommodating diverse textual inputs while maintaining the reasonability of the edited content as well as the specific skin details.

\begin{figure*}[ht]
  \centering
  \includegraphics[width=1.0\linewidth]{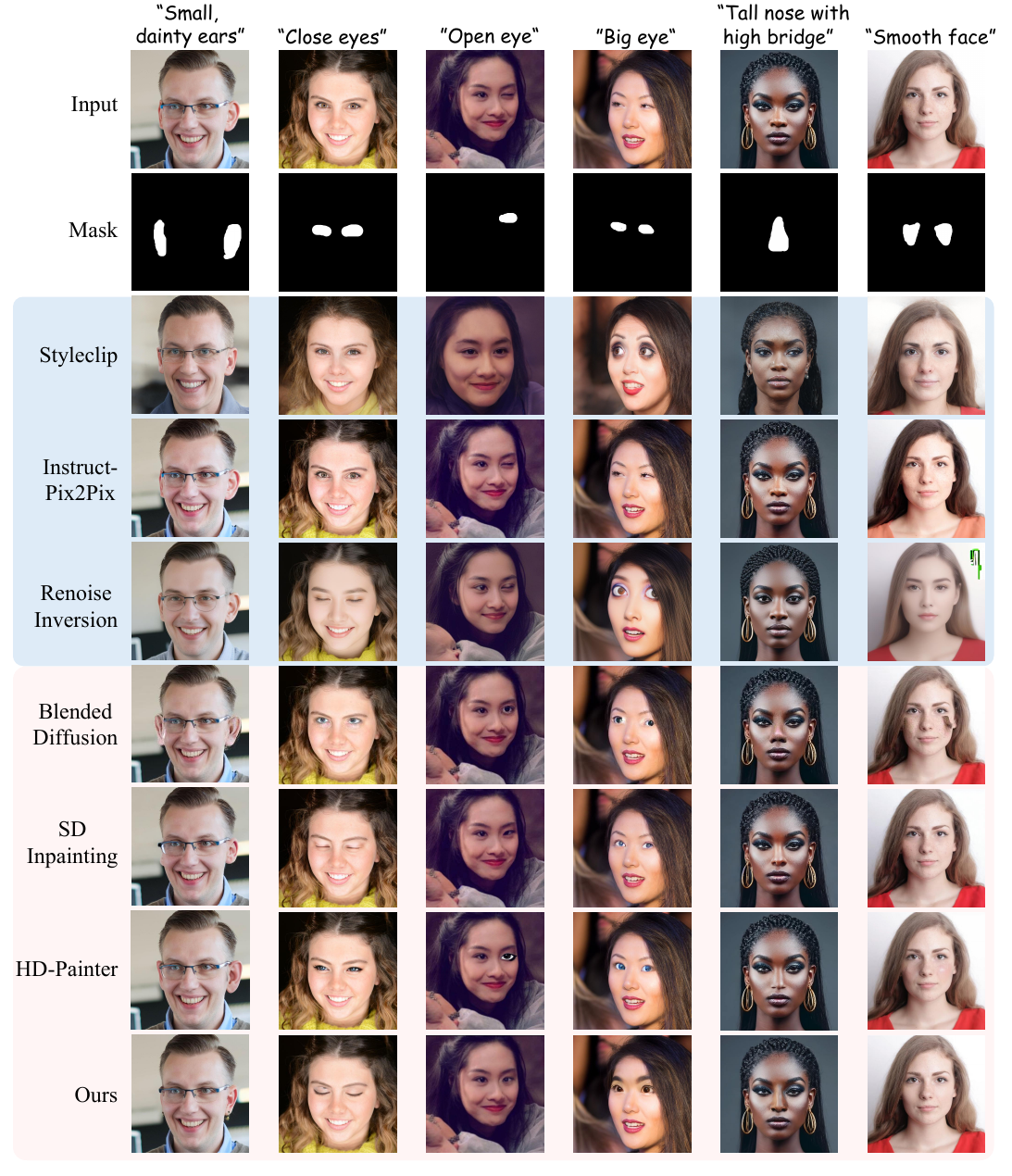}
  \caption{Comparison with related zero-shot methods. We extend our comparison to include a inpainting method \cite{manukyan2023hd} and an inversion-based method Renoise Inversion \cite{garibi2024renoise}. Other compared methods are InstructPix2Pix \cite{brooks2023instructpix2pix}, Blended Diffusion \cite{avrahami2022blended}, SD inpainting \cite{wang2022high} and  StyleClip \cite{patashnik2021styleclip}. 
  }
  \label{fig:extend1}
\end{figure*}

\begin{figure*}[ht]
  \centering
  \includegraphics[width=1.0\linewidth]{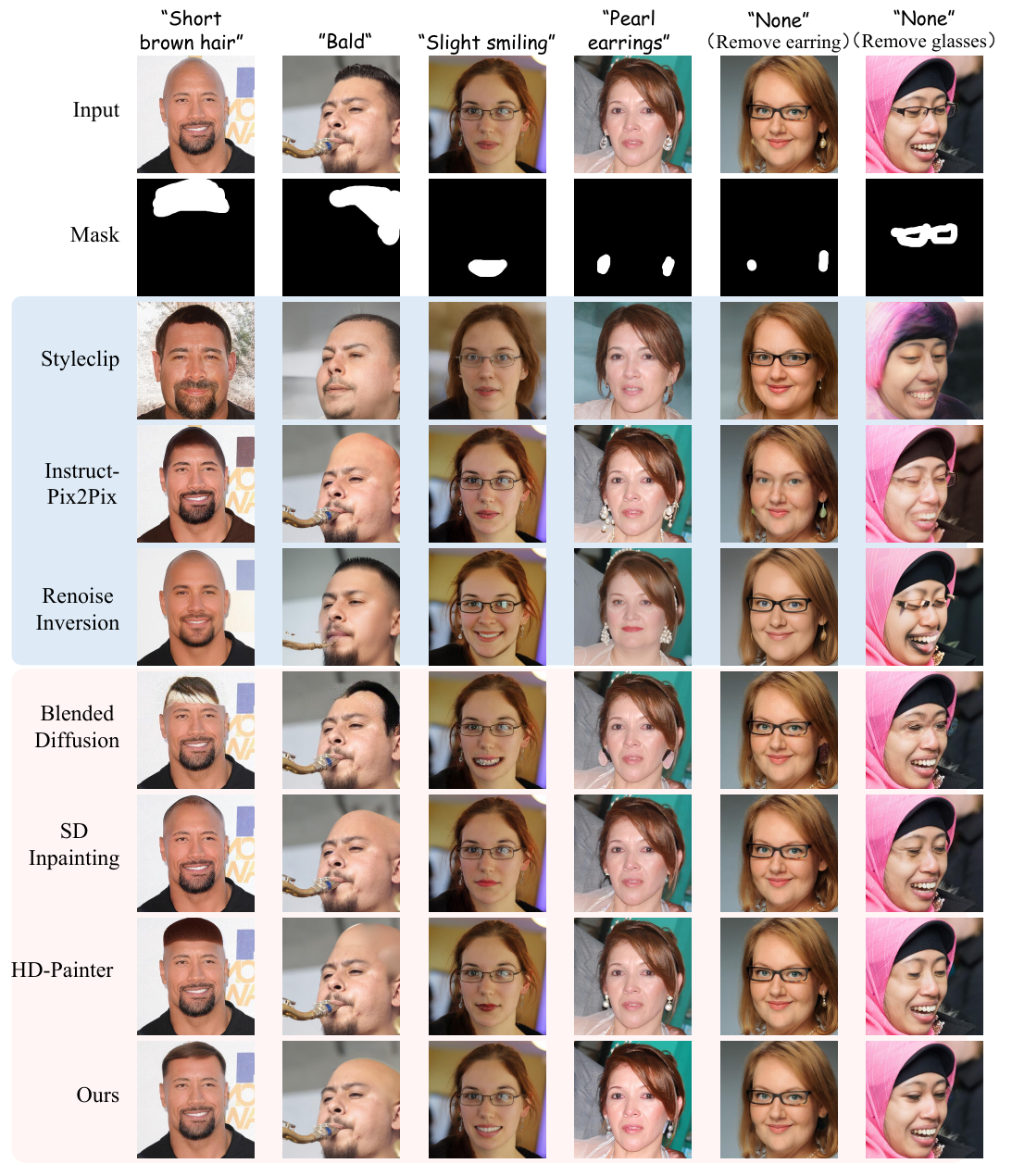}
  \caption{Comparison with related zero-shot methods. We extend our comparison to include a inpainting method \cite{manukyan2023hd} and an inversion-based method Renoise Inversion \cite{garibi2024renoise}. Other compared methods are InstructPix2Pix \cite{brooks2023instructpix2pix}, Blended Diffusion \cite{avrahami2022blended}, SD inpainting \cite{wang2022high} and  StyleClip \cite{patashnik2021styleclip}
  }. 
  \label{fig:extend2}
\end{figure*}

\textbf{Comparison with Existing Methods.} In Fig. \ref{fig:extend1} and Fig. \ref{fig:extend2}, we show additional visual comparison with image editing methods on more facial attributes.
In addition to the Qualitative Experiment Results in the main body, we include more inpainting methods (\cite{avrahami2022blended, manukyan2023hd}) and the Inversion-based method (Renoise Inversion \cite{garibi2024renoise} ) for the comparison. In these figures, we highlight the mask-free methods with blue color. 

Figs. \ref{fig:extend1} and \ref{fig:extend2} show that these compared approaches exhibit inferior performance when confronted with the task of editing local regions, due to the lack of mask integration.
Meanwhile, such methods often result in substantial leakage into incorrect regions during the process of localized editing with complex semantic textual description, or even changes of the individual identity (fourth column in Fig. \ref{fig:extend2}).

\begin{figure*}[!ht]
  \centering
  \includegraphics[width=1.0\linewidth]{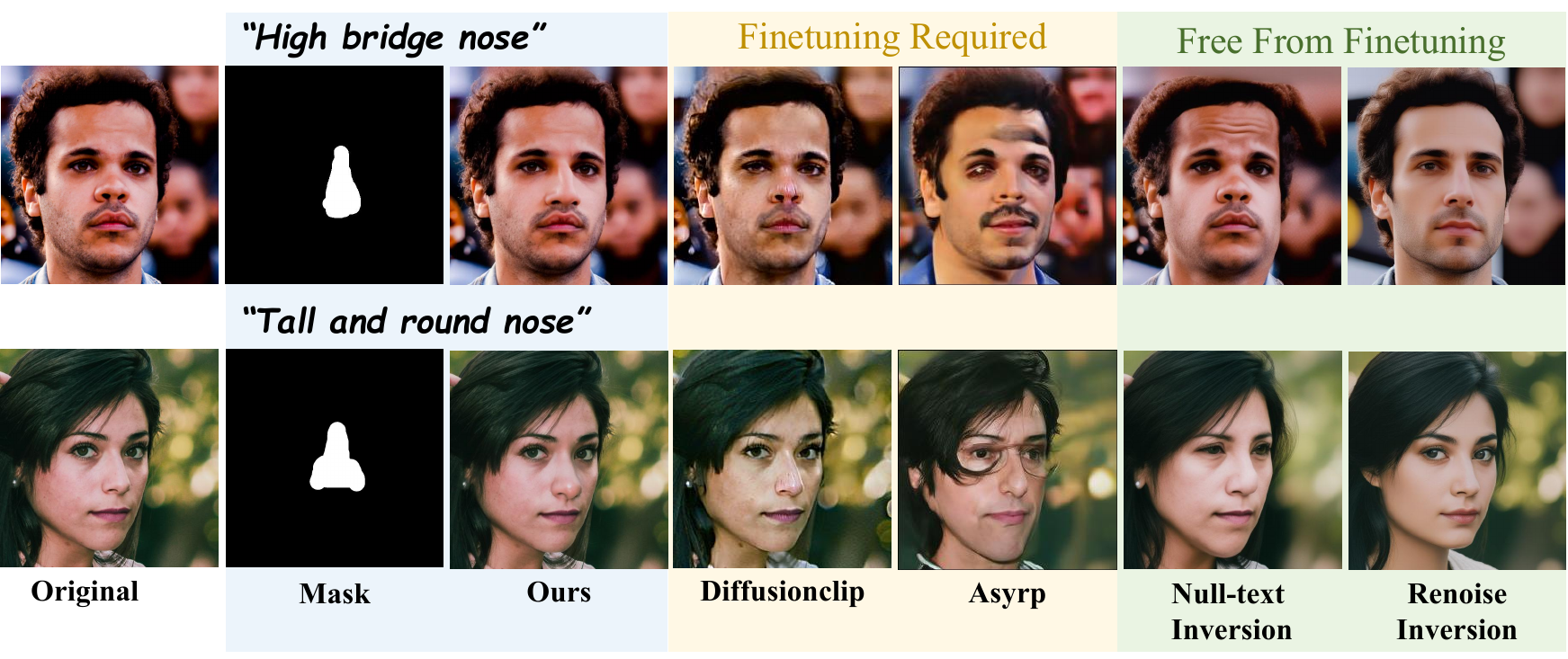}
  \caption{Comparison with the Inversion-based methods, i.e., Diffusionclip \cite{kim2022diffusionclip}, Asyrp \cite{KwonJU23}, Null-text Inversion \cite{mokady2023null} and Renoise Inversion \cite{garibi2024renoise}.}
  \label{fig:inversion}
\end{figure*}

\textbf{Comparison with Inversion-based Diffusion Methods.}
We also extend the comparison with existing approaches depending on inversion-based diffusion, including both the finetuning-required and finetuning-free paradigms. Among them, DiffusionClip \cite{kim2022diffusionclip} and Asyrp \cite{KwonJU23} both require additional finetuning for each previously unseen editing target with text prompt pairs. These inversion-based methods introduce a CLIP direction loss that aims to align the vector between the original and edited images with the one between the corresponding textual prompts in CLIP space. Null-text Inversion \cite{mokady2023null} and Renoise Inversion optimize the noise map during DDIM to further mitigate the error between the original image and the edited one in the resampling path during inference.
However, as illustrated in Fig. \ref{fig:inversion}, such methods fail to deal with localized editing, struggling to strike a trade-off between editability and fidelity. Specifically, they either perform minor editing on the target attributes or produce undesirable effects outside the target attributes. 

\begin{figure}[!ht]
  \centering
  \includegraphics[width=1.0\linewidth]{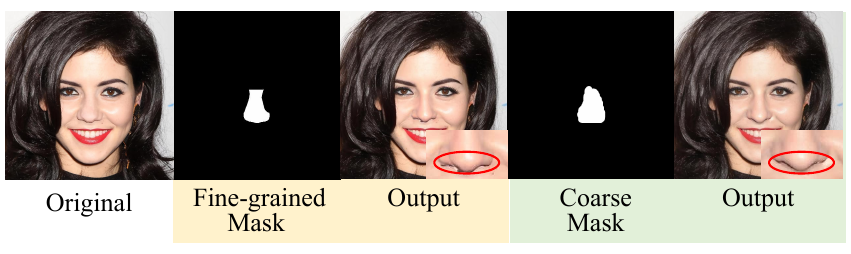}
  \caption{Comparison of the effects specific to coarse masks and fine-grained masks on the generated results.}
  \label{fig:fmask}
\end{figure}

\textbf{Coarse Masks v.s. Fine-grained Masks.}
To shed light on the reason behind using coarse masks, we conduct a toy experiment with an example to illustrate the effects of coarse and fine-grained masks on the generated results in Fig. \ref{fig:fmask}. It shows that fine-grained masks may lead to artifacts on the edges, and results in a noticeable boundary between the generated part and the unmasked regions during inference. However, this does not happen when the coarse masks are used.

\begin{table*}[!htbp]
\centering
\caption{Comparisons between the state-of-the-art methods and ours in terms of CLIP scores ( $\times 10^{2}$). ``CS Text" is the clip similarity between text input and the edited image, while ``CS Image" is the clip similarity between the original and edited images.
}
\label{tab:editing_performance}
\scalebox{0.95}{
\begin{tabular}{lcccccc}
\midrule
\multirow{2}{*}{\textbf{Method}} & \multicolumn{2}{c}{\textbf{FaRL \cite{zheng2021farl}}}& \multicolumn{2}{c}{\textbf{FLIP \cite{li2024flip}}} & \multicolumn{2}{c}{\textbf{CLIP \cite{2021radfordclip}}}  \\
\cmidrule(r){2-3}\cmidrule(r){4-5}\cmidrule(r){6-7}
& CS Text ($\uparrow$) & CS Image ($\uparrow$) & CS Text ($\uparrow$) & CS Image ($\uparrow$) & CS Text ($\uparrow$) & CS Image ($\uparrow$)\\ 
\midrule
SD Inpainting         & 21.21  & \textbf{93.98} & 18.52 &\textbf{93.26} &15.24 &\textbf{95.24}\\   
BrushNet              & 22.04  & 86.26 & 19.67 &83.41 &16.41 &88.38\\
Blended               & \underline{22.32}  & 87.35 & 20.15 &84.41 &\underline{16.45} &88.01\\
IntructPix2Pix        & 22.23  & 82.18 & \textbf{20.39} &82.18 &16.38 &87.64\\
\midrule
DiffusionClip   & 21.60  & 79.86 & 19.21 &80.13 &16.21 &84.63\\
Asyrp          & 19.72  & 72.90 & 16.71 &76.92 &15.88 &82.58\\
StyleClip       & 21.88  & 78.80 & 19.44 &80.47 &15.24 &83.28\\
\midrule
\rowcolor{gray!20} \textbf{Ours}  & \textbf{23.58} & \underline{91.16} &\underline{20.26} &\underline{90.07} &\textbf{16.88} &\underline{92.67}\\
\midrule
\end{tabular}
}
\end{table*}

\begin{table}[!htbp]
\centering
\footnotesize
\caption{Ablation study of our modules in terms of objective metrics.
}\label{tab:abla}
\scalebox{0.85}{
\begin{tabular}{lcccc}
\midrule
                &  FID/local-FID ($\downarrow$) & LPIPS ($\downarrow$) & MPS ($\uparrow$)   &HPSv2 ($\uparrow$)\\ 
                     \midrule
w/o CA$^{2}$             & 4.13       & 0.138       & 1.06    &  0.239  \\
Parallel Injection       & 9.80       & 0.153       & 1.03     &  0.262 \\
w/o STFG             & 5.94      & 0.097       & 1.09     & 0.264 \\
\rowcolor{gray!20} \textbf{Ours}     & \textbf{4.81}  & \textbf{0.085} & \textbf{/} & \textbf{0.264} \\
\midrule
\end{tabular}
}
\end{table}

\section{Extended Quantitative Results}

We incorporated CLIP's text score (CS Text) and image score (CS Image) as the Objective metrics. The former can reflect the consistency between the image and text, while the latter is used to evaluate the similarity between the original and the inpainted images, serving as a metric for image fidelity.
Since the target of our editing is specific to human faces, in addition to using the general CLIP \cite{2021radfordclip}, we also employ face-specific CLIP models (i.e., FaRL \cite{zheng2021farl}, Flip \cite{li2024flip}) as the base models for our evaluation.

The experimental results show that our method achieves the best CLIP text score under different CLIP models, indicating that our approach demonstrates better image-text consistency performance on the FFLEbench dataset containing complex textual descriptions.
At the same time, our method achieves the second-best CLIP image score, indicating that our approach achieves a relatively higher fidelity in terms of the overall edited image.
Notably, SD Inpainting exhibits overly high values on the CS Image metric, which is consistent with the LPIPS results reported in Tab. 1 of the main body, indicating that SD Inpainting may simply fill the mask while neglecting the prompts.

\section{Ablation Study with Quantitative Results}

The ablation study of qualitative experiments is conducted in the main body, we further present the quantitative results of the ablation study as shown in Tab. \ref{tab:abla}. The ``Parallel Injection" row represents the performance of the variant of our method after removing the spatial control of visual cross-attention, which may heavily rely on the visual cross-attention injection and thus impair the textual control capability.

\begin{figure}[!ht]
  \centering
  \includegraphics[width=1.0\linewidth]{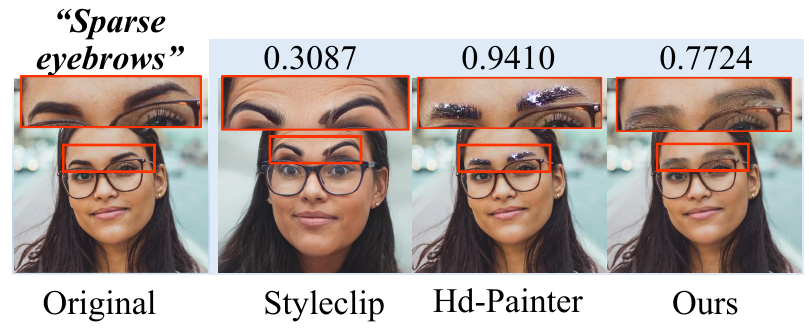}
  \caption{The limitation of identity (ID) similarity as a performance measure. Our result show better alignment with the textual prompt, despite lower ID similarity scores achieved.}
  \label{fig:id}
\end{figure}

\section{Discussion about the ID similarity metric} 

Since identity (ID) similarity serves as a vital metric in various face-related generation tasks, we study whether it is applicable in our local facial editing task.
Fig.\ref{fig:id} shows that ID similarity may not be as suitable for our task, i.e., the ID cues on the eyebrows are damaged although the target attribute is better aligned with the prompt. 
Therefore, while the ID similarity metric can reflect the fidelity of the edited image by measuring whether the ID is preserved, it may conflict with the goal of image editing.


\end{document}